\documentclass[review]{elsarticle}
\usepackage{lineno}
\usepackage{hyperref}
\usepackage{longtable}
\usepackage{amsmath,amssymb,amsfonts}
\usepackage{graphicx}
\usepackage{caption}
\usepackage{natbib}

\begin{document}

\begin{frontmatter}

\title{Improved Multi-label Classification with Frequent Label-set Mining and Association}

\author[1]{Anwesha Law}
\ead{anweshalaw\_r@isical.ac.in}

\author[1]{Ashish Ghosh\corref{correspondingauthor}}
\cortext[correspondingauthor]{Corresponding author}
\ead{ash@isical.ac.in}

\address[1]{Machine Intelligence Unit, Indian Statistical Institute, Kolkata, India}

\begin{abstract}
Multi-label (ML) data deals with multiple classes associated with individual samples at the same time. This leads to the co-occurrence of several classes repeatedly, which indicates some existing correlation among them. In this article, the correlation among classes has been explored to improve the classification performance of existing ML classifiers. A novel approach of frequent label-set mining has been proposed to extract these correlated classes from the label-sets of the data. Both co-presence (CP) and co-absence (CA) of classes have been taken into consideration. The rules mined from the ML data has been further used to incorporate class correlation information into existing ML classifiers. The soft scores generated by an ML classifier are modified through a novel approach using the CP-CA rules. A concept of certain and uncertain scores has been defined here, where the proposed method aims to improve the uncertain scores with the help of the certain scores and their corresponding CP-CA rules. This has been experimentally analysed on ten ML datasets for three ML existing classifiers which shows substantial improvement in their overall performance.
\end{abstract}

\begin{keyword}
Multi-label classification \sep frequent label-set mining \sep class correlation
\end{keyword}

\end{frontmatter}


\section{Introduction}
Shifting the traditional paradigm of machine learning from handling single-label data to multi-label data (MLD) is occurring at a fast pace \cite{herrera2016multilabel}. This is mostly due to the rise in the amounts of raw data available from the Internet which might be multi-label (ML) in nature. Here, every data sample can belong to more than one class simultaneously which divulges more information about the data than the single labels. Rather than forcefully assigning one label to a data sample, multi-label has the freedom to associate all classes related to a particular sample. It can be thought of as a generalization on single-label data.
Thus, the raw data need to be thoroughly analysed and utilized from an ML perspective for the benefit of users all over the world.
Multi-label learning can help to customize the user experience starting from software applications to physical real-world scenarios as well. Online social media platforms, news applications, audio/video streaming, e-commerce, etc. take into account the previously collected online activities of the user, to make their future experience more personal and based on their preferences. These give them an edge while holding on to the customers in the long run. Offline restaurants, retail stores, movie theatres, etc. also nowadays take into consideration user ratings, feedbacks, and other online information provided by the users to help the customers retain a good relationship and experience to improve their business. The concept of ML learning is being explored by researchers in various domains \cite{law2019multi, 9430169}.

Multi-label data in general is quite complex in nature.
Due to the simultaneous occurrence of multiple classes, the decision boundaries for multi-label classification invariably overlap. A common approach is to consider every class to be independent and perform prediction for each class separately. However, this loses out on any class correlation that might exist in the data. For example, if we are analysing user movie choices it can have genre classes like 'action', 'comedy', 'thriller', 'romance', 'science fiction (sci-fi)' etc. Among these, users preferring 'action' movies are highly likely to prefer 'thriller' as well. Movies genres are multi-label in nature hence not considering 'romantic-comedy' or 'action-thriller' 'action-sci-fi' as correlated classes might be a lack of judgement. This would eventually lose a lot of information, thus leading to deceiving results. Similar issues might be seen in other types of ML data as well.

In this article, an approach has been proposed to incorporate class correlations that exist in the ML data. This is done in order to improve the overall classification performance of any multi-label classifier that otherwise might have ignored these dependencies.
In this regard, a concept named ``frequent label-set mining'' (FLM) has been introduced. Here, the traditional concept of frequent itemset mining which finds the relation among frequently occurring itemsets has been utilized. Unlike the common approach of considering each feature as items, we have changed our perspective to a fresh angle, where the labels are considered as items. Thus, turning the label-sets of multi-label data into itemsets. This ensures that we find class correlations among frequently co-occurring labels when considered as frequent itemsets.
FLM is performed on the training data to extract the correlated classes. In this approach, concepts of co-presence (CP) and co-absence (CA) of classes have been presented. These help to extract relevant and irrelevant associations among classes in the form of association rules.
This FLM approach is then used in unison with a few benchmark ML classifiers to improve their classification accuracy.
The independent classifier trains itself on the training set and generates the soft classification scores. Depending on the ambiguity of the scores, they are marked as \textit{certain} or \textit{uncertain}. The \textit{uncertain} scores are then improved with the help of the important CP-CA rules extracted from the classes with \textit{certain} scores. The CP rules include frequently co-occurring classes, while the CA rules discard those classes which are frequently absent together. A novel score improvement formula has been defined to modify the soft classification scores.
Once the soft scores are updated depending on the class associations, they become more relevant or irrelevant and thus less ambiguous. After this, a hard classification is performed using a threshold function to generate the final label-sets.

The contribution of the proposed work is as follows.
\begin{itemize}
  \item Introduce the concept of ``frequent label-set mining'' for finding class/label correlations. This identifies co-presence (CP) and co-absence (CA) among classes to generate rules for relevant and irrelevant label-sets.
  \item Define a formula for the improvement of \textit{uncertain} scores incorporating CP-CA rules of \textit{certain} classes.
  \item Improve classification score for any classifier by incorporating class correlations with the help of frequent label-set mining.
\end{itemize}
The proposed algorithm, frequent label-set mining and association (FLMA), has been tested in combination with three ML classifiers to improve their performance on ten benchmark datasets. Experimental analysis indicates substantial improvement on the application of the proposed approach to incorporate class correlations on existing ML classifiers.

The rest of the article is organized as follows. Section \ref{Sec_RW} discusses some existing works in the field, and Section \ref{Sec_PW} elaborates on the proposed work. Section \ref{Sec_Results} includes all the experimental results done for this work and Section \ref{Sec_conc} concludes the article.

\section{Related Works}\label{Sec_RW}
Looking into the literature, there are various algorithms developed by researchers for efficient multi-label classification.
ML-KNN \cite{zhang2007ml} is a benchmark algorithm that is an adaptation of the traditional k-nearest neighbour algorithm, which incorporates first and second-degree neighbourhood information for ML classification.
ML-RBF \cite{zhang2009ml} is the ML variation of the radial basis function based neural network which also incorporates neighbourhood information.
Classifier Chains (CC) \cite{read2011classifier} use $C$ classifiers for $C$ classes while incorporating the class information along with the features to have a better classification.
It is an ensemble of binary classifiers which is dependent on the order of training of the classifiers. In its extended version, Ensemble of Classifier Chains (ECC) \cite{read2011classifier}, the authors aim to handle the drawbacks by introducing an ensemble model. These models aim to utilize the class information as well.
Random $k$ label-sets (RAKEL) \cite{tsoumakas2010random} is an ensemble method using several multi-class classifiers for various subsets of classes that acknowledge the label correlations in the data.
Multi-label classifier using Laplacian eigen map (MLLEM) \cite{kimura2016simultaneous} employs a non-linear embedding method which utilizes label-label, label-instance and instance-instance relations.

Apart from the benchmark methods, some existing works deal with association rule mining for multi-label data. Most researchers determine frequent itemsets and association rules based on the feature space of the data.
Authors in \cite{thabtah2004mmac} developed a multi-class, multi-label associative classification (MMAC) approach, which detects all frequent itemsets from sample features occurring in the data and the classes associated with them.
An application of the apriori algorithm has been done in \cite{qin2013application}. Here, the label-sets have been searched for correlations and the compound labels that have a strong association, have been replaced by single labels before classification. ML-KNN has been then used for multi-label classification, after which the labels have been reverted to their original form.


\section{Proposed Work}\label{Sec_PW}
The proposed system is designed to aid multi-label classifiers with frequent label-set mining to utilize class correlations. This aims to improve the classification performance achieved by a regular classifier that does not consider existing relations among classes. The following sections describe the major steps of the proposed method in detail.


\subsection{Frequent label-set mining}
Itemset mining is done to find frequently occurring items in a given dataset. In the scenario of multi-label data, each sample is associated with multiple classes at the same time. This set of relevant classes constitute the label-set for a particular sample. Similarly, the classes that are not associated with a particular sample, constitute its irrelevant label-set.
To map an MLD with itemset mining, each of the classes is considered as individual items, the label-sets are itemsets and then mining is performed in the output space. Frequent label-set mining would identify the classes that frequently occur together, hence can be considered to be correlated.
The aim of FLM is to generate positive rules for classes occurring together and negative rules for classes that are always absent together. This ensures that we cover all grounds for classes with a similar occurrence pattern.
Thus two concepts are introduced here:
\begin{enumerate}
  \item Co-presence (CP) label-sets: These indicate classes that frequently occur together. It helps to identify frequent relevant label-sets.
  \item Co-absence (CA) label-sets: These indicate classes that are frequently missing together. It helps to identify frequent irrelevant label-sets.
\end{enumerate}
FLM generates both CP and CA rules in this step.
However, if there are a large number of classes, the number of CP-CA label-sets would be huge and their corresponding rule generation might be computationally expensive.
For $C$ number of ML classes, the possible relevant and irrelevant label-sets is $2^C$. ML data tend to have a large number of classes, which in turn would create a large number of label-sets.
Hence, to reduce a bulk of the classes, some non-frequent classes are ignored at first to reduce the number of possible label-sets to some extent. This is an optional step taken for larger datasets. All classes with above-average occurrence frequency are considered for rule generation.
We aim to primarily handle classes that have the highest probability of occurrence.
From the reduced set of classes, the frequent label-sets are identified and their corresponding association rules are generated using the well-known FP-growth algorithm \cite{han2011data}. FP-growth builds a tree structure that is fast and can handle a large output dimension.

\subsubsection{FP-Growth Algorithm}
FP-growth is a popular method of mining frequent patterns (FP) in data that is specifically preferred over the Apriori algorithm since it can handle larger dimensional data. This is typically used on transactions related to a set of items to identify the frequently occurring itemsets from the entire list. The data can be accessed in two formats; the horizontal format, where the data is represented through transaction ids, and each id contains multiple items purchased in that transaction. The other is a vertical format, where each item is listed along with all the corresponding transactions in which they occur.

This algorithm has been adapted to be utilized in the proposed work to identify frequent label-sets in multi-label data. The data at hand is represented in a horizontal format, where each instance is associated with multiple classes which constitute its relevant label-set. Similarly, the classes that are not relevant to the instance constitute its irrelevant label-set. These relevant and irrelevant sets of classes are used to identify the frequent label-sets that exist in the data. For this purpose, the FP-growth algorithm has been employed on the label-sets of the multi-label data. The FP-growth algorithm first requires the building of the FP-Tree.
The steps to construct the FP-Tree for relevant label-sets are as follows:
\begin{enumerate}
  \item Scan the data and compute the support count for every individual label. The support of a class $Y$ is computed as,
  \begin{equation}\label{Eq_labelsupp}
    Support(Y) = \frac{\text{Samples with label}~Y}{N}.
  \end{equation}
  This can be done through one scan of the data.
  \item Sort the labels in descending order of support count. $min\_sup$ is the minimum support threshold which determines the frequent label-sets with support count above this threshold. All non-frequent labels can be eliminated.
  \item Begin creating the FP-Tree by forming an empty root node. Any node that gets added, has a \textit{support counter}.
  \item For each instance $i$,
  \begin{enumerate}[i]
    \item Sort its corresponding relevant labels in the order recorded in Step 2 and form a list $Rel_i$.
    \item Start at the root node, take labels from $Rel_i$ sequentially and traverse the tree.
    \item If a label exists in order of the sequence, increment its counter and traverse to the next branch.
    \item Add a branch to the tree, for every label in the sequence that does not exist and set its counter to one.
  \end{enumerate}
  \item Stop expanding the tree once all instances have been traversed.
\end{enumerate}
Once the FP-tree has been created, frequent itemsets can be extracted by traversing from the root node to the leaf nodes. The support counter at each node indicates the frequency of that particular path from the root. Using a minimum support threshold, the paths with high support count can be selected, thus identifying the frequent co-presence label-sets.
The FP-Tree for irrelevant label-sets is built and frequent co-absent label-sets are extracted in a similar way.

After the frequent label-sets have been identified association rules are generated from them. For every frequent label-set $\{Y_1, Y_2\}$, association rules $\{Y_1\}\rightarrow\{Y_2\}$ and $\{Y_2\}\rightarrow\{Y_1\}$ can be formed.
Each rule generated is of the form $\{Y_1\} \rightarrow \{Y_2\}$, where $\{Y_1\}$ and $\{Y_2\}$ can be single or multiple classes and the rule indicates an association between them. Each of these rules has two parameters associated with them as follows.
\begin{equation}\label{Eq_support}
  Support(\{Y_1\} \rightarrow \{Y_2\}) = \frac{\text{Samples with label-set} \{Y_1,Y_2\}}{N}
\end{equation}

\begin{equation}\label{Eq_conf}
  Confidence(\{Y_1\} \rightarrow \{Y_2\}) = \frac{\text{Samples with label-set} \{Y_1,Y_2\}}{\text{Samples with label} Y_1}
\end{equation}
These parameters indicate the importance of the rules, depending on which they have been used at a later stage. To select most important association rules, a minimum confidence threshold, $min\_conf$ is also set. All rules above the $min\_sup$ and $min\_conf$ thresholds constitute the final set of rules. The $min\_sup$ and $min\_conf$ thresholds for irrelevant label-sets is kept much higher since the frequency of 0's in the label space is higher than the frequency of  1's. This causes the support and confidence ranges for irrelevant labels to be higher than the relevant labels.
The rules generated from the relevant label-sets are the co-presence rules and the ones generated from the irrelevant label-sets are the co-absence rules.

Thus, the FLM step is performed in the training phase, and it helps to identify the relationship between sets of frequently occurring and frequently absent classes. It generates two sets of rules, one for relevant and one for irrelevant label-sets some of which are used later in the testing phase to improve the classification performance of a multi-label classifier.

\subsection{Soft classification}
In this step, an existing multi-label classifier is trained on the MLD in the training phase of the method to get the classification scores. Most classifiers do not explicitly consider class correlations or incorporate class information while training the classifier. Classifiers like ML-KNN, ML-RBF, multi-layer perceptrons, etc. are standard methods that can be improved using the proposed approach.
Final or hard classification is not performed in this step. The soft classification scores obtained from the ML classifier will be combined later with the rules generated in the FLM step. Since the MLC performs the training phase separately, it can be done in parallel while performing FLM.
In the testing phase, the trained model is used to predict classification scores, which are later improved using the proposed approach.

\subsection{Associative Correlation}
\begin{figure}[htbp]
  \centering
  \includegraphics[width=0.7\linewidth]{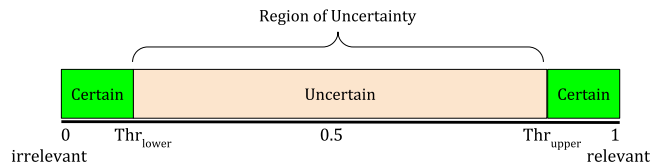}
  \caption{Region of Uncertainty}\label{Fig_certainty}
\end{figure}

This step combines the outputs generated from the FLM and soft classification steps.
The soft scores obtained from the classifier are aimed to be improved by correlating the CP-CA label-set rules generated in the first step. The scores can lie within the range of 0 to 1 or -1 to 1; here, we have considered [0,1]. As shown in \figurename ~\ref{Fig_certainty}, 0 indicates irrelevance while 1 indicates relevance.
The soft scores obtained can be segregated into two categories.
\begin{enumerate}
  \item \textbf{Certain scores} - The scores lying close to the boundary values are the ones the classifier is most certain about. Thus, classes that have scores lying close to 0 are surely irrelevant, and scores close to 1 are relevant.
  \item \textbf{Uncertain scores} - On the other hand, the region around the mean of the score range (in this case, 0.5) can be considered ambiguous. Thus, classification scores which lie close to the mean (0.5) are quite uncertain.
      A slight error by the classifier might 
      lead to misclassification.
\end{enumerate}
Thus, it can be said that the \textit{certain} scores do not need much intervention and can be considered to be correct. Most misclassifications occur due to the scores that lie in the uncertainty region.
In \figurename ~\ref{Fig_certainty}, a ``region of uncertainty'' has been demarcated. Two regions close to the boundaries (0 and 1) are considered to be \textit{certain}, while the region around the mean (here, 0.5) is \textit{uncertain}.
Boundary thresholds are determined which can distinguish the \textit{certain} regions from the \textit{uncertain} region. $Thr_{lower}$ determines the lower threshold closer to $0$, separating the \textit{certain} and \textit{uncertain} region for irrelevant classes, while $Thr_{upper}$ determines the upper threshold closer to $1$, separating the \textit{uncertain} and \textit{certain} scores for the relevant classes.
These thresholds are computed by fitting an S-membership function on the soft scores obtained from the classifier.
This work aims to improve the \textit{uncertain} scores with the help of the $certain$ scores and the relevant-irrelevant class correlations generated in the FLM step.

\subsubsection{Steps of Associative Correlation}
This phase combines the soft classification scores with the CP and CA rules.
\begin{enumerate}
  \item Assign hard labels to the \textit{certain} scores, i.e $score > Thr_{upper}$ and $score < Thr_{lower}$.
  \item Clean the overlapping CP and CA rules for the same set of classes. Keep the rules that have higher support and confidence.
  \item Sort the CP and CA rules in descending order of their confidence and support. Higher the confidence, more important is the rule.
  \item Sort the \textit{certain} classes and \textit{uncertain} classes based on their distance from the boundary. The shorter the distance from either boundary, the more confident is the score.
  \item Take a \textit{certain} class and find all the CP and CA rules involving that class.
  \item From the soft scores, find the most \textit{uncertain} class and improve its score.
  \item Score improvement: CP rule moves the score towards 1, CA rule moves the score towards 0.
\end{enumerate}


\subsubsection{Improving Uncertain Scores}
Using the \textit{certain} labels and the CP-CA rules, the scores of the \textit{uncertain} samples are improved.
With the help of CP rules, scores are increased to be closer to 1, and with CA rules, scores are moved towards 0. The chosen rule $\{X\} \rightarrow \{Y\}$ has \{X\} as a \textit{certain} label-set and \{Y\} as an \textit{uncertain} label-set. Equation \ref{Eq_new} determines the change $\Delta$ by which the \textit{uncertain} scores will be improved.
\begin{equation}\label{Eq_new}
  \Delta = Confidence(\{X\}\rightarrow \{Y\}).\frac{|Y_{score}-NB|}{|X_{score}-NB|},
\end{equation}
where, $NB$ is the nearest boundary.
The numerator incorporates the score of the \textit{uncertain} class ($Y_{score}$) and the confidence of the rule in question. Higher the confidence of the rule and the distance of $Y_{score}$ from the nearest boundary (i.e., a measure of ambiguity), the more the impact. The denominator determines the distance of $X_{score}$ (\textit{certain} class) from the nearest boundary. Better the \textit{certain} score, higher the impact.
$\Delta$ determines the shift from the original score to its desired score.

For CP rules, the scores are increased to move towards 1.
\begin{equation}\label{Eq1}
  Y_{score-new}=Y_{score-old}+\Delta.
\end{equation}

For CA rules, the score is decreased to move towards 0.
\begin{equation}\label{Eq2}
  Y_{score-new}=Y_{score-old}-\Delta.
\end{equation}

This computation weighs the confidence of the rule vs the ambiguity of the score.
For a particular \textit{uncertain} class, the new score is computed incorporating all relevant CP-CA rules. Addition and/or subtraction of $\Delta$ might occur several times for each rule applied. 

\subsection{Multi-label classification}
This is the final step. Once the changes are made by associative correlation, the new scores are recomputed from all the modifications that have been incorporated by the previous step. These scores are then converted to hard labels using a global threshold of 0.5.
While predicting labels for unseen data, first the ML classifier predicts scores for the test data. From these scores, the $certain$ and $uncertain$ scores are separated. The CP-CA rules from the training set are used for the $certain$ classes to improve the $uncertain$ scores.

\section{Results and Discussion}\label{Sec_Results}
Experimental analysis of the proposed work has been performed on ten standard ML datasets from Multi-label Classification Dataset Repository (http:// www.uco.es/kdis/mllresources/), namely, Emotions (music), Scene (image), Flags (image), Yeast (biology), Image (image), CHD\_49 (medicine), Yelp (web-text), Water quality (chemistry), Human\_PseAAC (biology) and GpositivePseAAC (biology). Seven performance metrics \cite{herrera2016multilabel} have been used for comparison, namely, Hamming loss (HL), ranking loss (RL), one error (OE), subset accuracy (SA), macro F1 (MacF1), micro F1 (MicF1) and accuracy (Acc).
Experiments have been done on a Windows 10 OS with core i7 processor on MATLAB 2015b.
The results are aggregated from multiple runs of 5-fold cross-validation.

\begin{figure}[htbp]
  \centering
  \includegraphics[width=\linewidth]{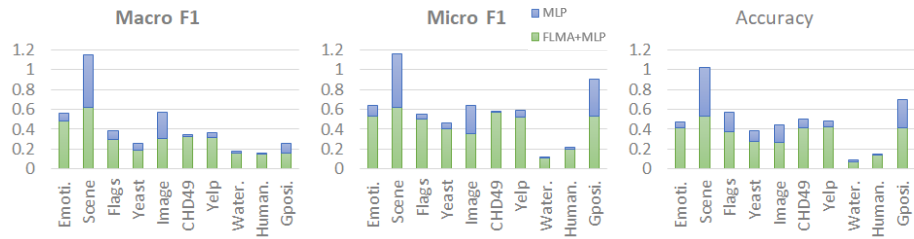}
  \caption{Improvement of MLP on application of FLMA}\label{Fig_MLPcompare}
\end{figure}

\begin{figure}[htbp]
  \centering
  \includegraphics[width=\linewidth]{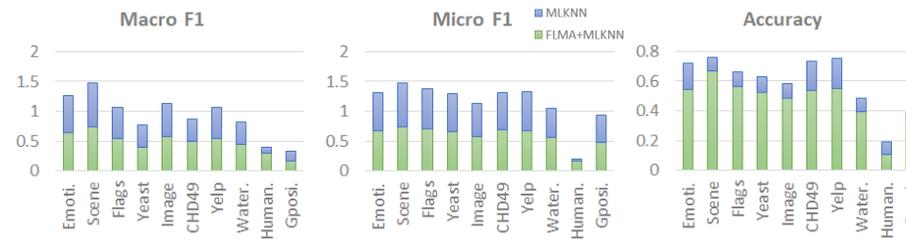}
  \caption{Improvement of ML-KNN on application of FLMA}\label{Fig_MLKNNcompare}
\end{figure}

\begin{figure}[htbp]
  \centering
  \includegraphics[width=\linewidth]{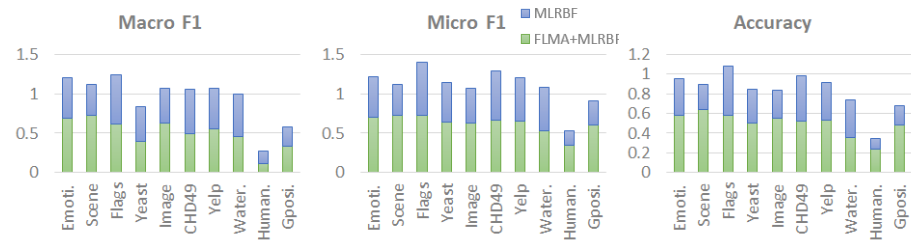}
  \caption{Improvement of ML-RBF on application of FLMA}\label{Fig_MLRBFcompare}
\end{figure}

The proposed model has been applied to benchmark ML classifiers to include label correlations information alongside classification. Figures \ref{Fig_MLPcompare} to \ref{Fig_MLRBFcompare} compare the performance of three algorithms namely, MLP, MLKNN and MLRBF respectively with and without the application of the proposed FLMA approach. The results for three metrics, macro F1, micro F1 and accuracy have been shown in the form of a stacked column chart for all ten datasets to analyse the impact of the proposed method. The coloured portions separately indicate the performance of each method. From the results, it is seen that the performances of the FLMA improved algorithms are significantly better than their stand-alone versions.

\begin{figure}[htbp]
  \centering
  \includegraphics[width=\linewidth]{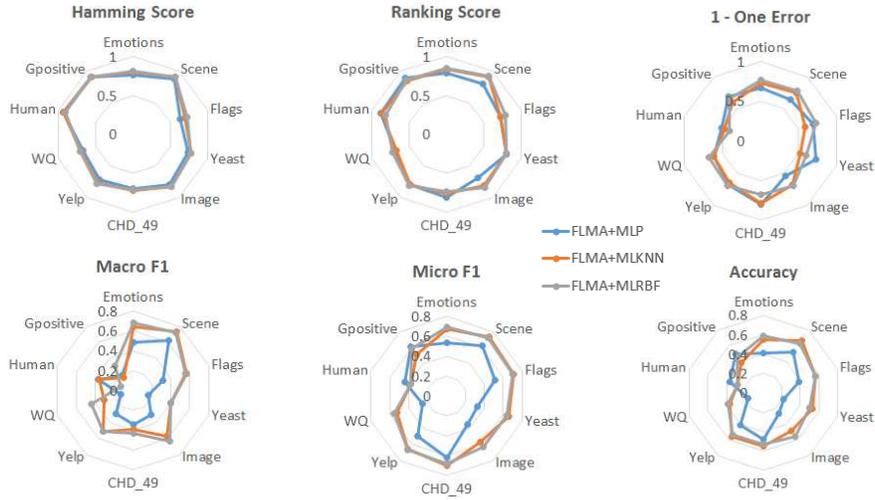}
  \caption{Comparison among the improved algorithms for ten datasets on six metrics}\label{Fig_LSMcompare}
\end{figure}

\figurename~ \ref{Fig_LSMcompare} shows the consolidated performance of the three FLMA improved algorithms for ten datasets and six metrics in the form of radar plots. The boundary covering a larger area on a radar plot depicts better performance. From the plots, it is seen that for all the six metrics, FLMA+MLKNN and FLMA+MLRBF perform very closely. However, the performance of FLMA+MLP is seen to falter for three metrics.

Table \ref{Tab_comp1} to \ref{Tab_comp10} compare the FLMA improved methods with four benchmark ML algorithms, namely RAKEL, CC, ECC and MLLEM. These are few well-known ML algorithms which utilize label information during classification. This makes these relevant to the proposed work and are thus used for comparative analysis. From the results, at a glance, it is seen that the FLMA improved algorithms performs better than the other ML classifiers in most cases.
Looking into specific performance metrics, it is seen that for the ranking-based one error metric, the outcome is lower than the other algorithms. However, the other ranking based metric, ranking loss seems to achieve best scores for the proposed approach.
Subset accuracy is a strict metric, which considers only those results which have completely correct labels. The proposed FLMA is seen to improve the SA metric quite well.
The other metrics like HL, MacF1, MicF1, Acc show improvement with the application of FLMA.
Among the three improved algorithms, FLMA+MLRBF achieves majority of the highest scores, followed by FLMA+MLKNN.
Overall, for all the ten datasets, the proposed FLMA is seen to improve the performance of the standard ML classifiers by incorporating class correlations that might have been otherwise ignored.


\begin{table}[htbp]
\centering
\caption{Comparison of methods for Emotions dataset}
\label{Tab_comp1}
\resizebox{0.8\textwidth}{!}{%
\begin{tabular}{lrrrrrrr}
\textbf{Method}      & \textbf{\textit{HL}} & \textbf{\textit{RL}} & \textbf{\textit{OE}} & \textbf{\textit{SA}} & \textbf{\textit{MacF1}} & \textbf{\textit{MicF1}} & \textbf{\textit{Acc}}\\
\textit{FLMA+MLP}     & 0.2339 & 0.2116 & 0.3305 & 0.1871 & 0.4835 & 0.5333 & 0.4114 \\
 \textit{FLMA+MLKNN}  & 0.1948 & 0.1615 & 0.2731 & 0.2969 & 0.6413 & 0.6717 & 0.5472 \\
  \textit{FLMA+MLRBF} & \textbf{0.1832} & \textbf{0.1479} & 0.2310 & \textbf{0.3306} & \textbf{0.6843} & \textbf{0.6964} & \textbf{0.5843} \\
  \textit{Rakel}      & 0.2369 & 0.6645 & 0.3052 & 0.2548 & 0.5686 & 0.6078 & 0.4837 \\
  \textit{CC}         & 0.2206 & 0.4852 & \textbf{0.1956} & 0.2464 & 0.5616 & 0.5964 & 0.4620  \\
  \textit{ECC}        & 0.2341 & 0.6290  & 0.2513 & 0.2649 & 0.5686 & 0.6149 & 0.4964 \\
  \textit{MLLEM}      & 0.2712 & 0.7752 & 0.3793 & 0.1383 & 0.4231 & 0.4327 & 0.3573
\end{tabular}}
\end{table}

\begin{table}[htbp]
\centering
\caption{Comparison of methods for Scene dataset}
\label{Tab_comp2}
\resizebox{0.8\textwidth}{!}{%
\begin{tabular}{lrrrrrrr}
\textbf{Method}      & \textbf{\textit{HL}} & \textbf{\textit{RL}} & \textbf{\textit{OE}} & \textbf{\textit{SA}} & \textbf{\textit{MacF1}} & \textbf{\textit{MicF1}} & \textbf{\textit{Acc}}  \\
 \textit{FLMA+MLP}   & 0.1107 & 0.1889 & 0.3436 & 0.5072 & 0.6187 & 0.6236 & 0.5308  \\
 \textit{FLMA+MLKNN} & \textbf{0.0872} & 0.0830 & 0.2414 & \textbf{0.6249}  & \textbf{0.7401} & \textbf{0.7338} & 0.6665  \\
 \textit{FLMA+MLRBF} & 0.0886 & \textbf{0.0750} & \textbf{0.2135} & 0.5933 & 0.7276 & 0.7211 & 0.6353  \\
 \textit{Rakel}      & 0.0991 & 0.7701 & 0.2443 & 0.6068 & 0.7246 & 0.7176 & \textbf{0.7008}  \\
 \textit{CC}         & 0.1072 & 0.6537 & 0.2575 & 0.5526 & 0.6974 & 0.6898 & 0.6230   \\
 \textit{ECC}        & 0.1062 & 0.7555 & 0.2202 & 0.5929 & 0.7135 & 0.7058 & 0.6685  \\
 \textit{MLLEM}      & 0.1093 & 0.9035 & 0.2908 & 0.6052 & 0.6984 & 0.6838 & 0.6771
\end{tabular}}
\end{table}

\begin{table}[htbp]
\centering
\caption{Comparison of methods for Flags dataset}
\label{Tab_comp3}
\resizebox{0.8\textwidth}{!}{%
\begin{tabular}{lrrrrrrr}
\textbf{Method}      & \textbf{\textit{HL}} & \textbf{\textit{RL}} & \textbf{\textit{OE}} & \textbf{\textit{SA}} & \textbf{\textit{MacF1}} & \textbf{\textit{MicF1}} & \textbf{\textit{Acc}}  \\
 \textit{FLMA+MLP}   & 0.3661 & 0.2413 & 0.2979 & 0.0617 & 0.2915 & 0.5035 & 0.3747 \\
 \textit{FLMA+MLKNN} & 0.2998 & 0.2527 & 0.3845 & \textbf{0.1184} & 0.5486 & 0.7004 & 0.5612 \\
 \textit{FLMA+MLRBF} & \textbf{0.2710}  & \textbf{0.2257} & 0.2912 & 0.0876 & \textbf{0.6107} & \textbf{0.7294} & \textbf{0.5836} \\
 \textit{Rakel}      & 0.2945 & 0.6632 & 0.0781 & 0.1136 & 0.4551 & 0.6739 & 0.5304 \\
 \textit{CC}         & 0.2932 & 0.5181  & 0.0382 & 0.0927 & 0.4834 & 0.6909 & 0.5409 \\
 \textit{ECC}        & 0.2911  & 0.5672 & \textbf{0.0363} & 0.1134 & 0.4838 & 0.6908 & 0.5439 \\
 \textit{MLLEM}      & 0.4685 & 0.7104 & 0.3073 & 0.0000 & 0.4095 & 0.4448 & 0.3211
 \end{tabular}}
\end{table}

\begin{table}[htbp]
\centering
\caption{Comparison of methods for Yeast dataset}
\label{Tab_comp4}
\resizebox{0.8\textwidth}{!}{%
\begin{tabular}{lrrrrrrr}
\textbf{Method}      & \textbf{\textit{HL}} & \textbf{\textit{RL}} & \textbf{\textit{OE}} & \textbf{\textit{SA}} & \textbf{\textit{MacF1}} & \textbf{\textit{MicF1}} & \textbf{\textit{Acc}}  \\
 \textit{FLMA+MLP}    & 0.2431 & \textbf{0.1876} & 0.2656 & 0.0372 & 0.1888 & 0.4021 & 0.2801   \\
  \textit{FLMA+MLKNN} & 0.2115 & 0.1909 & 0.4344 & \textbf{0.1498} & 0.3937 & \textbf{0.6545} & \textbf{0.5241} \\
  \textit{FLMA+MLRBF} & 0.2098 & 0.1925 & 0.3744 & 0.1351  & \textbf{0.4003} & 0.6458 & 0.5055 \\
  \textit{Rakel}      & \textbf{0.1996} & 0.6543 & 0.1489 & 0.1593 & 0.3644 & 0.6404 & 0.5106 \\
  \textit{CC}         & 0.2017 & 0.5538 & \textbf{0.1034} & 0.1422 & 0.3664 & 0.6389 & 0.5073 \\
  \textit{ECC}        & 0.2011 & 0.6371  & 0.1452 & 0.1447 & 0.3583 & 0.6376 & 0.5052 \\
  \textit{MLLEM}      & 0.3027 & 0.6801 & 0.5006 & 0.0095 & 0.1716 & 0.1909 & 0.1338
  \end{tabular}}
\end{table}

\begin{table}[htbp]
\centering
\caption{Comparison of methods for Image dataset}
\label{Tab_comp5}
\resizebox{0.8\textwidth}{!}{%
\begin{tabular}{lrrrrrrr}
\textbf{Method}      & \textbf{\textit{HL}} & \textbf{\textit{RL}} & \textbf{\textit{OE}} & \textbf{\textit{SA}} & \textbf{\textit{MacF1}} & \textbf{\textit{MicF1}} & \textbf{\textit{Acc}}  \\
\textit{FLMA+MLP}   & 0.2093 & 0.3126 & 0.4565 & 0.2369  & 0.3073 & 0.3549 & 0.2640  \\
\textit{FLMA+MLKNN} & 0.1714 & 0.1814 & 0.3235 & 0.4015 & 0.5747 & 0.5768 & 0.4840  \\
\textit{FLMA+MLRBF} & \textbf{0.1637} & \textbf{0.1595} & \textbf{0.2720}  & \textbf{0.4429}  & \textbf{0.6284} & \textbf{0.6274} & 0.5471 \\
\textit{Rakel}      & 0.1789  & 0.6842 & 0.3155 & 0.4095  & 0.6254 & 0.6213 & \textbf{0.5924} \\
\textit{CC}         & 0.1829 & 0.5362 & 0.2903  & 0.3971  & 0.5952 & 0.5936 & 0.5108 \\
\textit{ECC}        & 0.1819  & 0.6316 & 0.2805 & 0.4034  & 0.6150  & 0.6112 & 0.5531 \\
\textit{MLLEM}      & 0.1975  & 0.8043 & 0.3745 & 0.4155 & 0.5616 & 0.5595 & 0.5393
\end{tabular}}
\end{table}

\begin{table}[htbp]
\centering
\caption{Comparison of methods for CHD\_49 dataset}
\label{Tab_comp6}
\resizebox{0.8\textwidth}{!}{%
\begin{tabular}{lrrrrrrr}
\textbf{Method}      & \textbf{\textit{HL}} & \textbf{\textit{RL}} & \textbf{\textit{OE}} & \textbf{\textit{SA}} & \textbf{\textit{MacF1}} & \textbf{\textit{MicF1}} & \textbf{\textit{Acc}}  \\
\textit{FLMA+MLP}   & 0.3180 & 0.2590 & 0.3117 & 0.1006 & 0.3298 & 0.5669 & 0.4182 \\
\textit{FLMA+MLKNN} & 0.3158 & 0.2255 & 0.2754 & 0.1061 & 0.4933 & \textbf{0.6841} & \textbf{0.5354} \\
\textit{FLMA+MLRBF} & \textbf{0.2941} & \textbf{0.2176} & 0.2862 & \textbf{0.1431} & \textbf{0.4969} & 0.6676 & 0.5233 \\
\textit{Rakel}      & 0.3084 & 0.6261 & 0.1606 & 0.0952 & 0.4861 & 0.6436 & 0.5047 \\
\textit{CC}         & 0.2925 & 0.5289 & \textbf{0.0831} & 0.1112 & 0.4902 & 0.6609 & 0.5213 \\
\textit{ECC}        & 0.2922 & 0.6072 & 0.0992 & 0.1112 & 0.4791 & 0.6633 & 0.5273 \\
\textit{MLLEM}      & 0.4399 & 0.7159 & 0.4152 & 0.0054 & 0.0000 & 0.0000 & 0.0000
\end{tabular}}
\end{table}

\begin{table}[htbp]
\centering
\caption{Comparison of methods for Yelp dataset}
\label{Tab_comp7}
\resizebox{0.8\textwidth}{!}{%
\begin{tabular}{lrrrrrrr}
\textbf{Method}     & \textbf{\textit{HL}} & \textbf{\textit{RL}} & \textbf{\textit{OE}} & \textbf{\textit{SA}} & \textbf{\textit{MacF1}} & \textbf{\textit{MicF1}} & \textbf{\textit{Acc}}  \\
\textit{FLMA+MLP}   & 0.2723 & 0.2063 & 0.3519 & 0.2309 & 0.3147 & 0.5229 & 0.4285 \\
\textit{FLMA+MLKNN} & 0.2430 & 0.1892 & 0.3449 & 0.2362 & 0.5453 & \textbf{0.6675} & \textbf{0.5527} \\
\textit{FLMA+MLRBF} & 0.2193 & \textbf{0.1570} & 0.2731 & 0.2962 & \textbf{0.5569} & 0.6554 & 0.5272 \\
\textit{Rakel}      & \textbf{0.1663} & 0.7729 & \textbf{0.1879} & \textbf{0.4555} & 0.5455 & 0.6566 & 0.5471 \\
\textit{CC}         & 0.1805 & 0.6153 & 0.1432 & 0.3905 & 0.5211 & 0.6463 & 0.5221 \\
\textit{ECC}        & 0.1804 & 0.6476 & 0.1583 & 0.3882 & 0.5213 & 0.6460 & 0.5203 \\
\textit{MLLEM}      & 0.3275 & 0.6764 & 0.5555 & 0.0745 & 0.0095 & 0.0071 & 0.0032
\end{tabular}}
\end{table}

\begin{table}[htbp]
\centering
\caption{Comparison of methods for Water Quality dataset}
\label{Tab_comp8}
\resizebox{0.8\textwidth}{!}{%
\begin{tabular}{lrrrrrrr}
\textbf{Method}      & \textbf{\textit{HL}} & \textbf{\textit{RL}} & \textbf{\textit{OE}} & \textbf{\textit{SA}} & \textbf{\textit{MacF1}} & \textbf{\textit{MicF1}} & \textbf{\textit{Acc}}  \\
\textit{FLMA+MLP}   & 0.3526 & 0.3120 & 0.3434 & 0.0019 & 0.0624 & 0.1141 & 0.0704 \\
\textit{FLMA+MLKNN} & 0.3149 & 0.3001 & 0.4924 & \textbf{0.0152} & 0.4382 & \textbf{0.5649} & \textbf{0.3927} \\
\textit{FLMA+MLRBF} & \textbf{0.3074} & \textbf{0.2668} & 0.3824 & 0.0104 & \textbf{0.4602} & 0.5309 & 0.3569 \\
\textit{Rakel}      & 0.3113 & 0.4955 & 0.2019 & 0.0145 & 0.3581 & 0.4599 & 0.2964 \\
\textit{CC}         & 0.3192 & 0.3486 & \textbf{0.1009} & 0.0149 & 0.3822 & 0.4939 & 0.3264 \\
\textit{ECC}        & 0.3101 & 0.4269 & 0.1726 & 0.0150 & 0.3767 & 0.4838 & 0.3187 \\
\textit{MLLEM}      & 0.4144 & 0.6751 & 0.4566 & 0.0047 & 0.1036 & 0.1058 & 0.0701
\end{tabular}}
\end{table}

\begin{table}[htbp]
\centering
\caption{Comparison of methods for Human\_PseAAC dataset}
\label{Tab_comp9}
\resizebox{0.8\textwidth}{!}{%
\begin{tabular}{lrrrrrrr}
\textbf{Method}      & \textbf{\textit{HL}} & \textbf{\textit{RL}} & \textbf{\textit{OE}} & \textbf{\textit{SA}} & \textbf{\textit{MacF1}} & \textbf{\textit{MicF1}} & \textbf{\textit{Acc}}  \\
\textit{FLMA+MLP}   & \textbf{0.0840} & 0.1780 & 0.6135 & 0.1224 & 0.0456 & 0.2013 & 0.1347 \\
\textit{FLMA+MLKNN} & 0.0870 & 0.1839 & 0.6668 & 0.0888 & 0.0299 & 0.1604 & 0.1041 \\
\textit{FLMA+MLRBF} & 0.0913 & \textbf{0.1779} & 0.6017 & \textbf{0.1674} & \textbf{0.1159} & \textbf{0.3431} & \textbf{0.2429} \\
\textit{Rakel}      & 0.0993 & 0.3396 & 0.4674 & 0.1606 & 0.0935 & 0.3245 & 0.2326 \\
\textit{CC}         & 0.0996 & 0.2389 & \textbf{0.3267} & 0.1484 & 0.0853 & 0.2811 & 0.2050 \\
\textit{ECC}        & 0.1027 & 0.3071 & 0.4282 & 0.1555 & 0.0936 & 0.3015 & 0.2377 \\
\textit{MLLEM}      & 0.1182 & 0.7521 & 0.7349 & 0.2135 & 0.1069 & 0.2425 & 0.2384
\end{tabular}}
\end{table}

\begin{table}[htbp]
\centering
\caption{Comparison of methods for GPositivePseAAC dataset}
\label{Tab_comp10}
\resizebox{0.8\textwidth}{!}{%
\begin{tabular}{lrrrrrrr}
\textbf{Method}      & \textbf{\textit{HL}} & \textbf{\textit{RL}} & \textbf{\textit{OE}} & \textbf{\textit{SA}} & \textbf{\textit{MacF1}} & \textbf{\textit{MicF1}} & \textbf{\textit{Acc}}  \\
\textit{FLMA+MLP}   & 0.0926 & 0.1416 & 0.4030 & 0.4109 & 0.1557 & 0.5318 & 0.4184 \\
\textit{FLMA+MLKNN} & 0.0979 & 0.1438 & 0.4303 & 0.3283 & 0.1662 & 0.4695 & 0.3384 \\
\textit{FLMA+MLRBF} & \textbf{0.0826} & \textbf{0.1106} & \textbf{0.3326} & \textbf{0.4669} & 0.3316 & \textbf{0.6035} & \textbf{0.4847} \\
\textit{Rakel}      & 0.0971 & 0.6898 & 0.3634 & 0.4297 & 0.3565 & 0.5905 & 0.4196 \\
\textit{CC}         & 0.1022 & 0.6059 & 0.3478 & 0.3928 & 0.3689 & 0.5513 & 0.3630 \\
\textit{ECC}        & 0.9979 & 0.0154 & 0.0134 & 0.0138 & 0.0080 & 0.0118 & 0.0134 \\
\textit{MLLEM}      & 0.0873 & 0.9068 & 0.3462 & 0.4501 & \textbf{0.4109} & 0.5887 & 0.4570
\end{tabular}
}
\end{table}

\section{Conclusion}\label{Sec_conc}
This article proposes a frequent label-set mining technique as an additional improvement on any existing multi-label classifier that does not specifically consider the correlation between classes. For ML data, multiple classes constitute frequent label-sets which indicates an association between these labels. A novel approach of frequent label-set mining for ML data has been proposed which generates co-present and co-absent rules. The proposed method then improves the classification scores of an existing ML classifier by incorporating these rules. The proposed method has been tested on ten ML datasets with seven performance measures in combination with three ML classifiers. The results indicate substantial improvement on the application of the proposed technique with respect to the existing ML classifiers.

\section*{Acknowledgement}
This work of Anwesha Law was supported by the Indian Statistical Institute, India and Technology Innovation Hub on Data Science, Big Data Analytics and Data Curation under Grant NMICPS/006/MD/2020-21 dt 16.10.2020.

\bibliography{References}
\end{document}